**Title:**

# ModernBERT is More Efficient than Conventional BERT for Chest CT Findings Classification in Japanese Radiology Reports

**Authors:**

Yosuke Yamagishi, MD, MSc[1]

Tomohiro Kikuchi, MD, PhD, MPH[2, 3]

Shouhei Hanaoka, MD, PhD[1, 4]

Takeharu Yoshikawa, MD, PhD[2]

Osamu Abe, MD, PhD[1,4]

**Affiliations**

1. Division of Radiology and Biomedical Engineering, Graduate School of Medicine, The University of Tokyo, Tokyo, Japan
2. Department of Computational Diagnostic Radiology and Preventive Medicine, The University of Tokyo Hospital, Tokyo, Japan
3. Department of Radiology, School of Medicine, Jichi Medical University, Shimotsuke, Tochigi, Japan
4. Department of Radiology, The University of Tokyo Hospital, Tokyo, Japan

**Corresponding Author**

Yosuke Yamagishi, MD, MSc

Division of Radiology and Biomedical Engineering, Graduate School of Medicine, The University of Tokyo

7-3-1 Hongo, Bunkyo-ku, Tokyo, 113-8655, Japan

Email: yamagishi-yosuke0115@g.ecc.u-tokyo.ac.jp

**Abstract**

Japanese language models for medical text classification face challenges with complex vocabulary and linguistic structures in radiology reports. This study compared three Japanese models—BERT Base, JMedRoBERTa, and ModernBERT—for multi-label classification of 18 chest CT findings. Using the CT-RATE-JPN dataset, all models were fine-tuned under identical conditions. ModernBERT showed clear efficiency advantages, producing substantially fewer tokens and achieving faster training and inference than the other models while maintaining comparable performance on the internal test dataset (exact match accuracy: 74.7% vs. 72.7% for BERT Base). To assess generalizability, we additionally constructed RR-Findings, an external dataset of 243 naturally written Japanese radiology reports annotated using the same schema. Under this domain-shifted setting, performance differences became pronounced: BERT Base outperformed both JMedRoBERTa and ModernBERT, whereas ModernBERT showed the largest decline in exact match accuracy. Average precision differences were smaller, indicating that ModernBERT retained reasonable ranking ability despite reduced calibration. Overall, ModernBERT offers substantial computational efficiency and strong in-domain performance but remains sensitive to real-world linguistic variability. These results highlight the need for more diverse natural-language training data and domain-specific calibration strategies to improve robustness when deploying modern transformer models in heterogeneous clinical environments.

**Introduction**

In recent years, the rapid development of large language models (LLMs) has demonstrated impressive capabilities in understanding radiological texts [1,2], showcasing remarkable potential for medical text analysis. Nevertheless, in the field of Japanese radiology reports analysis, Bidirectional Encoder Representations from Transformers (BERT) continues to be the established standard for natural language processing tasks [3], consistently demonstrating reliable performance in analyzing radiological reports and clinical documentation [4–7]. BERT, with just 110 million parameters in its base configuration, offers significant processing efficiency compared to recent LLMs which are tens to thousands of times larger. Numerous studies have applied BERT to various Japanese radiology report tasks, such as detecting actionable reports [8], extracting clinical information [9], determining the presence of malignant findings [10], and scoring the importance of head CT findings [11].

For specialized medical applications, these task-specific small-scale language models remain preferential despite the capabilities of larger models, for several critical reasons. Domain-adapted small models often achieve comparable performance on specific tasks after fine-tuning with relevant data [12,13]. In our previous research [14], we conducted finding classification of chest CT radiology reports using the BERT model and achieved performance comparable to OpenAI's GPT-4o, which was considered state-of-the-art at that time. Their smaller parameter count enables more efficient training and faster inference, significantly reducing computational costs when processing thousands of documents. Most importantly, the use of small-scale models that can be deployed locally addresses the privacy concerns inherent in medical contexts—unlike cloud-based LLMs such as OpenAI's ChatGPT [15] or Anthropic's Claude [16] that require data transmission to external servers. This local deployment capability ensures sensitive medical data remains within secure hospital infrastructure, a crucial advantage for clinical applications.

In this context, the recent introduction of ModernBERT presents a further advancement that may challenge this established standard [17]. Within this landscape of specialized small-scale models, conventional BERT architectures have adequately served radiology applications, yet they exhibit inherent inefficiencies when processing Japanese medical text. The recently released Japanese version of ModernBERT offers significant improvements in tokenization efficiency, enabling more effective processing of Japanese medical terminology and syntax [18]. ModernBERT's optimized tokenizer, coupled with architectural refinements, promises faster training speeds despite having comparable parameter counts to traditional BERT models.

Beyond efficiency improvements, ModernBERT incorporates architectural innovations and has undergone extensive pre-training on vast datasets, suggesting potential performance advantages in downstream tasks. These enhancements theoretically position ModernBERT to outperform conventional BERT models in specialized domains such as medical text analysis, though this advantage requires empirical verification in medical-specific applications.

This study evaluates the performance differential between conventional BERT and ModernBERT when applied to the classification of findings in radiological reports. By applying both models to this clinical task, we aim to quantify any performance advantages offered by ModernBERT and assess whether its theoretical benefits translate to practical improvements in radiological text analysis.

## Methods

### Ethical considerations

This research was conducted as a retrospective study utilizing the open-access dataset available for research purposes. As we exclusively used de-identified data from this research dataset without accessing any protected patient information, ethical approval was deemed unnecessary. Given that this study is based on a previously published open-access dataset and does not contain any studies with human participants performed by any of the authors, informed consent was not required.

### Dataset

The CT-RATE-JPN dataset (https://huggingface.co/datasets/YYama0/CT-RATE-JPN) [14], which was developed in our previous research, is a Japanese-translated version of radiology reports from the original CT-RATE dataset (https://huggingface.co/datasets/ibrahimhamamci/CT-RATE) [19–21]. The CT-RATE-JPN dataset was specifically created to facilitate Japanese medical AI model development and evaluation. We selected this dataset for our comparative study as it represents the only publicly available Japanese radiology text corpus with sufficient training data to effectively train deep learning models and with annotated test data suitable for rigorous evaluation. The dataset consists of 22,778 training reports and 150 test reports. For this study, we divided the training data in a 4 : 1 ratio (18,222 : 4,556) for training and validation, and used the 150 test reports to evaluate model performance, as illustrated in Figure 1. The dataset contains structured annotations for 18 different medical conditions and findings commonly observed in chest CT examinations, as shown in Supplementary Table S1. For our experiments, we used the findings sections of the reports, as they contain the detailed descriptions of radiological observations relevant to our classification tasks.

Additionally, because no publicly available Japanese dataset contained external evaluation data aligned with CT-RATE-style multi-label chest CT findings, we newly constructed an independent evaluation dataset named RR-Findings (https://huggingface.co/datasets/YYama0/RR-Findings). RR-Findings is a labeled set of 243 Japanese chest CT radiology reports derived from the publicly released MedNLP-SC Radiology Report TNM Staging (RR-TNM) Dataset [22]. Each report was annotated with 18 clinically relevant chest CT findings following the schema of the CT-RATE dataset. Each report was initially annotated by a postgraduate year (PGY) 5 radiology resident. A PGY 12 board-certified radiologist then reviewed all annotations, proposed necessary corrections, and final labels were established through collaborative consensus. This two-step annotation workflow ensured both the clinical validity and the linguistic fidelity of the labels. Importantly, because RR-TNM consists of real, naturally written Japanese radiology reports authored by nine board-certified radiologists and covering 27 lung cancer cases, RR-Findings provides an evaluation resource that reflects genuine clinical writing style, terminology usage, and descriptive variability—complementing CT-RATE-JPN, which is based on translated reports. Thus, RR-Findings serves as a reliable, expert-verified external evaluation dataset for assessing model generalization to authentic Japanese radiology reporting.

**Model architecture**

We compared three Japanese language models based on transformer architecture for chest CT findings classification, as illustrated in Figure 2.

For the conventional BERT implementation, we utilized the pre-trained "tohoku-nlp/bert-base-japanese-v3" model available from HuggingFace. This model was trained on the CC-100 corpus for 1 million steps followed by additional training on the Wikipedia corpus for another 1 million steps. The final fine-tuned model using CT-RATE-JPN is identical to our previously released "YYama0/CT-BERT-JPN" from our prior research [14]. This model demonstrated superior performance compared to GPT-4o in 11 out of 18 conditions.

For the clinically oriented RoBERTa-based model [23], we employed "alabnii/jmedroberta-base-manbyo-wordpiece," which was pre-trained on large-scale Japanese medical corpora and optimized using the Manbyo disease-name dictionary [24]. This model incorporates a medical domain–specific vocabulary designed to better handle clinical terminology commonly observed in radiology reports.

For ModernBERT, we used the "sbintuitions/modernbert-ja-130m" model, which introduces several architectural refinements. This model underwent a multi-phase training approach: initial pre-training with 3.51 trillion tokens from Japanese and English web corpora using a 1,024 token sequence length, followed by two context extension phases. The first extension phase used 430 billion tokens of high-quality Japanese and English data with an 8,192 token sequence length and best-fit packing. The final phase incorporated 450 billion tokens of high-quality Japanese data (150 billion tokens trained for 3 epochs) without sequence packing. The results are detailed in Table 1.

**Tokenization approaches**

Tokenization is the process of breaking text into smaller units (tokens) that language models can process. For clinical text analysis, effective tokenization is crucial for capturing medical terminology accurately. This process significantly influences how models interpret specialized vocabulary, abbreviations, and complex medical terms commonly found in radiological reports.

The tokenization strategies differ between the three models. BERT Base utilizes a standard vocabulary size of 32,768. JMedRoBERTa employs a medical domain–optimized WordPiece vocabulary constructed using the Manbyo dictionary, allowing improved handling of disease names and specialized clinical terminology. In contrast, ModernBERT employs an enhanced tokenization approach with a significantly larger vocabulary size of 102,400 including English vocabulary. These improvements in tokenization potentially enhance the model's ability to process multilingual medical content and specialized terminology commonly found in Japanese clinical documents.

**Model Training**

All models were implemented using the Transformers library (version 4.48.3) to ensure consistency in implementation and allow for direct comparison. The models were fine-tuned with identical hyperparameters: learning rate of 2e-5, batch size of 8, 4 training epochs, and weight decay of 0.01. Evaluation was performed at the end of each epoch, with model checkpoints saved accordingly. The best-performing model checkpoint based on evaluation metrics was selected as the final model for our experiments. The models were trained to minimize the binary cross-entropy loss function for the multi-label classification task.

All experiments were conducted on Google Colaboratory using an NVIDIA L4 GPU with 24GB VRAM.

**Evaluation and statistical analysis**

We evaluated all three models using multiple metrics to provide a comprehensive performance comparison.

For computational efficiency, we measured training runtime, reports processed per second, and steps per second. Additionally, to isolate the impact of model architecture from tokenization differences, we conducted controlled experiments where both models were trained with identical fixed token lengths (64, 128, 256, and 512 tokens). These experiments measured training and inference samples processed per second across different sequence lengths, allowing for direct performance comparison independent of each model's native tokenization approach.

Moreover, token statistics—such as average tokens, median values, standard deviation, and token count range—were analyzed to assess differences in tokenization efficiency across the three models. Because JMedRoBERTa exhibited intermediate tokenization characteristics between ModernBERT and BERT Base, it was included in the descriptive token analysis, while efficiency experiments were conducted using ModernBERT and BERT Base.

Performance on CT findings classification was primarily assessed using F1 scores for the detection of 18 distinct medical conditions from the radiological reports. In addition to this per-condition evaluation, overall performance was also measured using an exact match accuracy metric, in which a test report was considered correct only if all 18 labels were accurately predicted. This holistic metric allowed us to capture the models' ability to fully represent the clinical findings. To statistically compare the overall performance among the three models, we first applied Cochran's Q test to evaluate whether any significant differences existed in exact match accuracy. When the global test indicated significance, pairwise comparisons were subsequently conducted using McNemar's test with Bonferroni correction to control for multiple testing.

For the external evaluation, we additionally considered the characteristics of the dataset, which consists of lung cancer staging reports and therefore exhibits substantial class imbalance. Several CT findings included in our task are typically absent or rarely mentioned in such reports, resulting in highly skewed label distributions. Because F1 scores can become unstable under extreme imbalance, especially for rare findings, we complemented the per-label F1 assessment by computing average precision (AP) for each finding as well as the macro-averaged mean AP. AP was selected not only for its robustness under severe label imbalance but also because it has been adopted in representative long-tail medical imaging benchmarks, such as the CXR-LT challenge, which similarly evaluate performance under highly imbalanced label distributions [25,26]. These additional metrics therefore provide a more reliable and informative characterization of model performance under the severely imbalanced conditions of the external dataset.

## Results

### Tokenization efficiency

Among the three models evaluated (BERT Base, JMedRoBERTa, and ModernBERT), ModernBERT demonstrated the most efficient tokenization when processing radiological reports in the internal test dataset. Across all 150 reports, ModernBERT consistently produced fewer tokens than BERT Base. JMedRoBERTa also reduced token counts compared with BERT Base, with results that consistently fell between those of ModernBERT and BERT Base. The average tokens per document were 258.1 for ModernBERT, 289.7 for JMedRoBERTa, and 339.6 for BERT Base. Similarly, the median token count was lower with ModernBERT (251.0 vs. 278.0 for JMedRoBERTa and 327.5 for BERT Base). ModernBERT's tokenization was also more consistent, shown by a lower standard deviation (69.4 vs. 79.1 for JMedRoBERTa and 90.1 for BERT Base). The maximum token count was 523 with ModernBERT, compared with 579 for JMedRoBERTa and 683 for BERT Base. As shown in Figure 3 (ModernBERT vs. BERT Base), the histogram reveals ModernBERT's distribution is shifted toward lower token counts, with a higher concentration around the mean compared to BERT Base's more dispersed distribution. The distribution for JMedRoBERTa lies between the two, reflecting moderate improvements in tokenization efficiency. These findings confirm ModernBERT's superior tokenization efficiency across the entire distribution of radiological reports. The detailed results are presented in Table 2.

Given that BERT Base is a widely adopted baseline in radiology NLP, the following analysis focuses on comparisons between ModernBERT and BERT Base. ModernBERT achieved a 32.8% reduction in overall token count, from 198 tokens to 133 tokens in the best-performing example from the complete text analysis provided in Supplementary Figure S1. As detailed in Supplementary Figure S2, this improvement is highlighted by examples where ModernBERT achieves significant reductions in token count compared to BERT. Notably, the term for main bronchi is tokenized as 2 tokens in ModernBERT, whereas BERT splits it into 4 tokens. The phrase meaning "was not observed" is processed as 2 tokens in ModernBERT, while BERT breaks it into 6 tokens. Additionally, ModernBERT's enhanced English vocabulary leads to better handling of mixed-language medical phrases like "crazy paving appearance", which is tokenized as 4 tokens in ModernBERT, compared to BERT's tokenization requiring 10 tokens. Through these examples, we can see that efficient tokenization is possible for anatomical and commonly used medical expressions, as well as mixed-language expressions frequently found in Japanese medical texts.

**Training and inference efficiency**

As BERT Base is a standard baseline, the following analysis compares ModernBERT mainly with BERT Base.

The data shows that ModernBERT offers significant performance improvements over BERT Base in both training and inference tasks. As illustrated in Figure 4, ModernBERT achieves a 1.65× speedup during training (processing 38.82 samples per second compared to BERT Base's 23.58) and a 1.66× speedup during inference (139.90 samples per second versus 84.18).

These efficiency gains are substantial when considering practical applications in clinical settings or other contexts where computational resources are limited or expensive. ModernBERT completed the full training cycle in approximately 1873 seconds, faster than both BERT Base (3154 seconds) and JMedRoBERTa (2892 seconds), making it the most resource-efficient option among the three models.

As illustrated in Supplementary Figure S3, ModernBERT demonstrates faster convergence compared to BERT Base, with its training loss decreasing more rapidly during the initial training steps. Similarly, ModernBERT shows quicker improvement in validation loss and achieves higher F1 scores more rapidly in the first epoch. This indicates ModernBERT's ability to reach better performance with less training epochs. However, despite ModernBERT's faster initial convergence, BERT Base ultimately achieves a slightly better validation loss by the end of training.

When comparing performance across fixed token lengths, we observed an interesting pattern in computational efficiency. As shown in Supplementary Figure S4, at the shortest token length (64), BERT Base actually outperformed ModernBERT in processing speed for both training and inference tasks. However, as token length increased (128, 256, and 512), ModernBERT progressively demonstrated superior efficiency, with the performance gap widening at longer sequences. This trend became particularly pronounced at the 512 token length, where ModernBERT showed significantly faster processing speeds. These results suggest that ModernBERT's architectural improvements provide substantial computational advantages for longer sequences - a characteristic especially valuable for processing comprehensive radiology reports that typically contain extensive clinical observations.

**Classification performance on the internal test dataset**

When evaluating the models' performance on identifying 18 different medical conditions in radiological reports, we observed nuanced differences across categories among BERT Base, JMedRoBERTa, and ModernBERT. ModernBERT achieved the highest F1 scores in several findings, including arterial wall calcification (1.0000), emphysema (0.9688), pleural effusion (0.9744), bronchiectasis (0.9524), and atelectasis (1.0000). JMedRoBERTa showed the best or tied-best performance in multiple categories such as medical material (0.9333), coronary artery wall calcification (0.9888), lung nodule (0.9753), and consolidation (0.8302). BERT Base achieved the top performance in selected findings including pericardial effusion (1.0000) and pulmonary fibrotic sequela (0.9247), and matched the highest score in several others. The detailed per-finding results are summarized in Table 3.

In addition to the per-finding analysis, overall performance was assessed using an exact match metric, where all 18 labels must be correctly predicted for a report to be counted as correct. BERT Base achieved 109 exact matches out of 150 reports (0.7267 accuracy), ModernBERT achieved 112 exact matches (0.7467 accuracy), and JMedRoBERTa achieved 108 exact matches (0.7200 accuracy). Cochran's Q test indicated no statistically significant difference among the three models ($Q = 1.0400$, $p = 0.5945$), suggesting that their overall exact match performance is comparable.

Across all findings, performance differences among the three models remained modest, generally within a ±3% range, indicating that ModernBERT's efficiency advantages—such as reduced tokenization overhead and faster training—did not compromise classification accuracy. Consistently high AP values in Supplementary Table S2 further confirm the strong overall performance of all three models.

**Classification performance on the external test dataset**

Table 4 summarizes the per-finding sample counts and prevalence within the RR-Findings dataset. Three findings—Hiatal hernia, Mosaic attenuation pattern, and Medical material—had zero positive samples ($N = 0$) and were therefore excluded from the external evaluation to avoid undefined F1 scores and unstable statistical estimates.

Analyzing errors on an external, independently authored dataset is essential for understanding model generalizability beyond translated or standardized corpora. On the RR-Findings dataset, overall performance showed a clear reversal compared with the internal evaluation. Exact match accuracy was highest for BERT Base (0.6996), followed by JMedRoBERTa (0.6255), and lowest for ModernBERT (0.5514). Cochran's Q test indicated a significant overall difference among the three models ($Q = 21.6000$, $p = 2.04 \times 10^{-5}$).

Pairwise comparisons using McNemar's test with Bonferroni correction ($\alpha = 0.0167$) revealed that BERT Base significantly outperformed ModernBERT ($p = 6.10 \times 10^{-6}$), whereas the differences between BERT Base and JMedRoBERTa ($p = 0.0461$) and between ModernBERT and JMedRoBERTa ($p = 0.1331$) were not statistically significant after correction.

Across individual findings, F1 and AP results (Table 4 and Supplementary Table S3) exhibited a heterogeneous but largely consistent pattern. BERT Base and JMedRoBERTa demonstrated stable performance on common findings such as lymphadenopathy, lung opacity, and pleural effusion, achieving F1 scores in the 0.88–1.00 range and AP values exceeding 0.95. JMedRoBERTa achieved the highest F1 and AP for consolidation (F1 = 0.7368; AP = 0.8748), whereas ModernBERT showed strong performance in findings such as bronchiectasis and emphysema, with AP values comparable to the other models. In contrast, all models showed reduced F1 and AP for several less frequent conditions, though these differences were less interpretable due to limited sample sizes. When aggregated across all findings, the models showed measurable differences in mAP (0.8119 for BERT Base, 0.8066 for JMedRoBERTa, and 0.7776 for ModernBERT); however, the degree of separation was smaller than that observed in exact match accuracy, reflecting more moderate variation in their ranking-based performance.

Error analysis on the external test dataset

The most substantial performance drop was observed for pulmonary fibrotic sequela, where ModernBERT's F1 score decreased to 0.0833. A detailed review of the RR-Findings reports revealed that this category was frequently expressed using paraphrased terminology such as *"chronic inflammatory change"*, rather than the explicit wording found in CT-RATE-JPN. This lexical variation resulted in reduced performance across all models, with the most pronounced degradation in ModernBERT. Notably, however, the corresponding AP values (0.6748, 0.6236, 0.6346) remained comparatively similar across models, indicating that the primary source of degradation was threshold sensitivity rather than a fundamental loss of discriminative ability.

A similar pattern was observed for lung nodule, where ModernBERT yielded a lower F1 score (0.8291) compared with BERT Base (0.8879) and JMedRoBERTa (0.9237). In the external dataset, radiologists frequently used abbreviations such as *"GGN (ground-glass density nodule)"* in place of explicit descriptors, contributing to reduced detection performance.

**Discussion**

This study evaluated three Japanese transformer-based language models—BERT Base, JMedRoBERTa, and ModernBERT—for chest CT findings classification in radiology reports. Across our experiments, ModernBERT provided substantial efficiency gains in both training and inference while achieving performance on the internal test dataset that was comparable to, or slightly better than, the other two models. However, on the externally constructed RR-Findings dataset, the performance gains of ModernBERT were limited, and BERT Base ultimately achieved the highest exact match accuracy. These results suggest that, despite its clear computational advantages, ModernBERT remains more sensitive to domain shift than conventional BERT and a medically specialized model such as JMedRoBERTa.

The tokenization analysis demonstrated that ModernBERT required substantially fewer tokens per document than the other models, resulting in clear efficiency advantages. These benefits translated into strong performance within the internal test dataset, where ModernBERT achieved accuracy comparable to BERT Base and JMedRoBERTa, indicating robust fine-tuning behavior under in-domain conditions.

However, its performance on the external RR-Findings dataset was more limited. ModernBERT showed a marked decrease in exact match accuracy, whereas its average precision values, although lower than those of the other two models, exhibited only moderate separation. This pattern suggests that while the model preserved reasonable ranking ability for individual findings, its output confidence distribution was more sensitive to domain shift. These findings highlight that additional strategies—such as incorporating more diverse natural Japanese texts during fine-tuning or optimizing decision thresholds in new clinical domains—may be necessary to improve ModernBERT's robustness to unseen data.

While ModernBERT showed sensitivity to domain shift, its efficiency improvements highlight a promising direction for model development. Traditionally, many Japanese radiology report studies have relied on BERT models [6,8,10,11,27–31], which, while effective, place a considerable computational burden on training and inference processes. ModernBERT, however, offers the potential to significantly reduce this burden by processing text more efficiently. While architectural improvements contribute to this efficiency, the optimization of tokenization methods has particularly led to a substantial reduction in required computational resources. In addition to improving processing speed, it is important to note that the use of GPUs for model training requires significant power consumption, raising concerns about environmental impact [32]. As computational resources become increasingly strained, optimizing model efficiency not only accelerates the processing of large volumes of radiology reports but also contributes to sustainability by reducing energy consumption. Moreover, efficient finding extraction has practical relevance beyond model development. Structured information derived from radiology reports is essential for retrospective case review, epidemiologic research, and the construction of new machine learning models. Improvements in computational efficiency directly accelerate these iterative processes, enabling faster dataset curation and model refinement and ultimately supporting smoother integration of automated finding extraction into clinical and research workflows.

The extended sequence length capability of ModernBERT (8,192 tokens vs. 512 in conventional BERT) represents a significant advantage for medical document analysis, although this feature was not fully utilized in our evaluation dataset where documents were relatively short. Future research should investigate ModernBERT's performance on longer medical documents, such as comprehensive case reports or full radiology studies, where this extended capacity would be more beneficial. In particular, ModernBERT's extended context window enables the simultaneous processing of multiple clinical documents, creating opportunities for multimodal clinical information integration. Combining radiology reports with patient medical history, laboratory results, and treatment records could provide a more comprehensive view of patient conditions and potentially improve diagnostic accuracy. For example, recent multimodal datasets such as INSPECT, which integrate CT imaging, report text, and structured EHR data [33], highlight opportunities where long-sequence models like ModernBERT could be more fully utilized. Developing similar resources in the Japanese clinical domain would enable future evaluations of extended-context modeling.

This study has several limitations. Although we strengthened the evaluation by incorporating the newly constructed RR-Findings dataset in addition to CT-RATE-JPN, both resources consist solely of chest CT reports, and radiology reporting also spans MRI and other anatomical regions; broader modalities should therefore be examined to assess generalizability. Even with this addition, the total number of evaluation samples remained modest (393 reports in total), limiting statistical robustness. ModernBERT's performance degradation under domain-shifted conditions further highlights the need for more diverse, naturally written Japanese radiology corpora during fine-tuning, as well as domain-aware recalibration strategies. Finally, while ModernBERT supports substantially extended sequence lengths, our datasets consisted of relatively short reports, preventing a full assessment of its long-document capabilities—an aspect that should be explored in future work.

In conclusion, ModernBERT demonstrates promising results for Japanese radiology reports analysis, offering substantial efficiency improvements while maintaining performance compared to conventional BERT models. At the same time, its reduced performance under domain-shifted conditions highlights the need for further improvements in generalization. These findings suggest that architectural and tokenization enhancements can meaningfully improve the applicability of transformer-based models to specialized medical text tasks, but also underscore the importance of robustness when deployed across heterogeneous clinical environments. Future work should evaluate ModernBERT across a wider range of imaging modalities and document types, including longer and naturally written reports, and investigate fine-tuning and calibration strategies that enhance resilience to domain shift and better capture the variability of real-world radiological language.

**Acknowledgment**

The Department of Computational Diagnostic Radiology and Preventive Medicine, The University of Tokyo Hospital, is sponsored by HIMEDIC Inc. and Siemens K.K. There is no funding for this research.

**Competing interests**

Authors TK and TY belong to The Department of Computational Diagnostic Radiology and Preventive Medicine at The University of Tokyo Hospital, which is sponsored by HIMEDIC, Inc. and Siemens Japan KK.

**Data Availability**

The datasets and models used in this study are publicly available. The CT-RATE-JPN dataset, which was used for training and evaluation, can be accessed at https://huggingface.co/datasets/YYama0/CT-RATE-JPN. The external test dataset constructed in this study, RR-Findings, is also available at https://huggingface.co/datasets/YYama0/RR-Findings.

The fine-tuned models used in this work, including BERT Base (https://huggingface.co/YYama0/CT-BERT-JPN), ModernBERT (https://huggingface.co/YYama0/CT-ModernBERT-JPN), and JMedRoBERTa (https://huggingface.co/YYama0/CT-JMedRoBERTa), are likewise publicly accessible.

**Author contributions**

All authors contributed to the conceptualization of the study. YY implemented the model, conducted the evaluation, performed statistical analysis, and wrote the initial draft of the manuscript. TK contributed to statistical analysis and editing of the manuscript. SH, TY, and OA contributed to manuscript editing and supervision of the project. All authors reviewed and approved the final version of the manuscript.

**Tables:**

| Model | BERT Base | JMedRoBERTa | ModernBERT |
|---|---|---|---|
| Base Model | tohoku-nlp/bert-base-japanese-v3 | alabnii/jmedroberta-base-manbyo-wordpiece | sbintuitions/modernbert-ja-130m |
| Parameters | 111 million | 109 million | 132 million |
| Pretraining | 1M steps on CC-100 & Wikipedia each | Japanese medical corpus (disease-name optimized Manbyo dictionary) | 4.4 trillion tokens in 3 phases on Japanese and English corpus |
| Sequence length | 512 | 512 | 8192 |

Table 1: Comparison of key parameters across three Japanese language models: BERT Base, JMedRoBERTa, and ModernBERT. JMedRoBERTa incorporates a disease-name–optimized WordPiece vocabulary trained on a Japanese medical corpus, whereas ModernBERT was trained on substantially larger multilingual data and supports sequences up to 16× longer, enabling more efficient processing of lengthy clinical documents.

| Model | Mean token count | Median token count | Standard deviation | Minimum token count | Maximum token count |
|---|---|---|---|---|---|
| BERT Base | 339.6 | 327.5 | 90.4 | 160 | 683 |
| JMedRoBERTa | 289.7 | 278.0 | 79.1 | 139 | 579 |
| ModernBERT | 258.1 | 251.0 | 69.6 | 119 | 523 |

Table 2: Detailed tokenization statistics comparing BERT Base, JMedRoBERTa, and ModernBERT on the internal test dataset. The table summarizes key metrics including mean, median, standard deviation, and the range of token counts for each model, allowing a quantitative assessment of tokenization efficiency and sequence length characteristics across the three architectures.

| Findings | BERT Base | JMedRoBERTa | ModernBERT | N (%) |
|---|---|---|---|---|
| Medical material | 0.8750 | 0.9333 | 0.8667 | 14 (9.3%) |
| Arterial wall calcification | 0.9800 | 0.9703 | 1.0000 | 49 (32.7%) |
| Cardiomegaly | 0.9583 | 0.9362 | 0.9583 | 25 (16.7%) |
| Pericardial effusion | 1.0000 | 0.9565 | 0.9565 | 12 (8.0%) |
| Coronary artery wall calcification | 0.9778 | 0.9888 | 0.9778 | 45 (30.0%) |
| Hiatal hernia | 1.0000 | 1.0000 | 1.0000 | 24 (16.0%) |
| Lymphadenopathy | 0.9730 | 0.9730 | 0.9730 | 37 (24.7%) |
| Emphysema | 0.9524 | 0.9677 | 0.9688 | 31 (20.7%) |
| Atelectasis | 0.9899 | 0.9800 | 1.0000 | 49 (32.7%) |
| Lung nodule | 0.9693 | 0.9753 | 0.9753 | 82 (54.7%) |
| Lung opacity | 0.9369 | 0.9273 | 0.9455 | 55 (36.7%) |
| Pulmonary fibrotic sequela | 0.9247 | 0.9231 | 0.8966 | 47 (31.3%) |
| Pleural effusion | 0.9500 | 0.9500 | 0.9744 | 19 (12.7%) |
| Mosaic attenuation pattern | 1.0000 | 1.0000 | 1.0000 | 25 (16.7%) |
| Peribronchial thickening | 0.8333 | 0.8333 | 0.8421 | 21 (14.0%) |
| Consolidation | 0.8276 | 0.8302 | 0.8070 | 24 (16.0%) |
| Bronchiectasis | 0.9302 | 0.9302 | 0.9524 | 20 (13.3%) |
| Interlobular septal thickening | 0.9333 | 0.9333 | 0.9333 | 7 (4.7%) |

Table 3: Comparison of F1 scores across BERT Base, JMedRoBERTa, and ModernBERT for classification of 18 medical findings in chest CT reports on the internal test dataset. The N (%) column indicates the number and proportion of positive samples for each finding.

| Findings | BERT Base | JMedRoBERTa | ModernBERT | N (%) |
|---|---|---|---|---|
| Lymphadenopathy | 0.9324 | 0.8745 | 0.9429 | 148 (60.9%) |
| Lung nodule | 0.8879 | 0.9237 | 0.8291 | 121 (49.8%) |
| Pleural effusion | 1.0000 | 1.0000 | 0.9781 | 67 (27.6%) |
| Lung opacity | 0.9440 | 0.8777 | 0.9206 | 62 (25.5%) |
| Emphysema | 0.9748 | 0.9431 | 0.9431 | 59 (24.3%) |
| Atelectasis | 0.9697 | 0.9697 | 0.9388 | 51 (21.0%) |
| Pulmonary fibrotic sequela | 0.5882 | 0.2308 | 0.0833 | 23 (9.5%) |
| Consolidation | 0.7083 | 0.7368 | 0.5667 | 17 (7.0%) |
| Interlobular septal thickening | 1.0000 | 0.9412 | 0.8889 | 8 (3.3%) |
| Pericardial effusion | 0.8889 | 0.8571 | 1.0000 | 4 (1.6%) |
| Cardiomegaly | 0.0000 | 0.0000 | 0.0000 | 4 (1.6%) |
| Arterial wall calcification | 0.4000 | 0.5000 | 0.4000 | 1 (0.4%) |
| Bronchiectasis | 0.6667 | 0.6667 | 1.0000 | 1 (0.4%) |
| Coronary artery wall calcification | 0.6667 | 0.6667 | 0.6667 | 1 (0.4%) |
| Peribronchial thickening | 0.0000 | 0.0000 | 0.0000 | 1 (0.4%) |
| Hiatal hernia | NA | NA | NA | 0 (0.0%) |
| Mosaic attenuation pattern | NA | NA | NA | 0 (0.0%) |
| Medical material | NA | NA | NA | 0 (0.0%) |

Table 4: Comparison of F1 scores across BERT Base, JMedRoBERTa, and ModernBERT for classification of 18 medical findings in chest CT reports on the external test dataset. Due to domain shift, several findings have zero positive samples in this dataset, and their F1 scores are reported as NA. Findings are ordered in descending order of sample size according to the N (%) column.

**Figure legends:**

Figure 1: Overview of the CT-RATE-JPN dataset and data splitting approach. A total of 22,778 reports were used for training and validation (split 4:1 into 18,222 training reports and 4,556 validation reports), and an existing set of 150 reports served as the original test dataset.

Figure 2: Overview of the chest CT findings classification framework. The process begins with the extraction of finding sections from a large dataset of radiology reports (n = 22,778). These text data are then used to fine-tune two Japanese language models (BERT Base and ModernBERT) for chest CT findings classification task.

Figure 3: Comparison of token length distributions between ModernBERT and BERT Base on radiological reports of test data. Left: Box plot showing the lower median and narrower distribution for ModernBERT compared to BERT Base. Right: Histogram with density curves illustrating ModernBERT's more efficient tokenization, with consistently lower token counts across the dataset. KDE: Kernel Density Estimation.

Figure 4: A bar chart comparing the training and inference speeds (in samples per second) of BERT Base and ModernBERT. ModernBERT demonstrates a 1.65× speedup during training and a 1.66× speedup during inference compared to BERT Base.

**Figures:**

Training Data Distribution

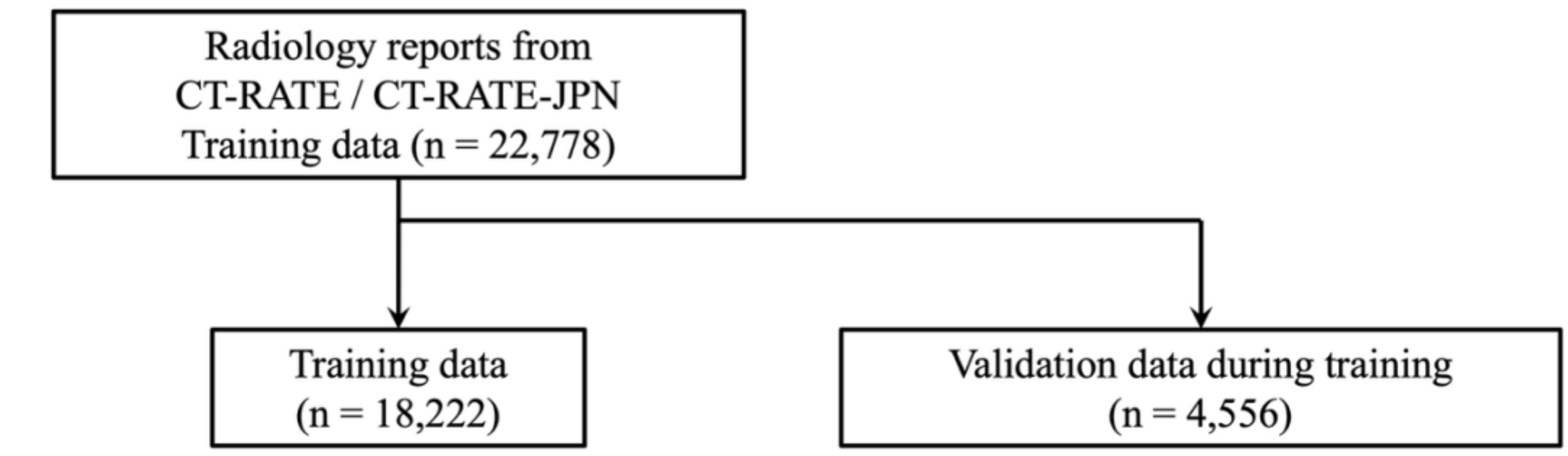

Test data for performance comparison

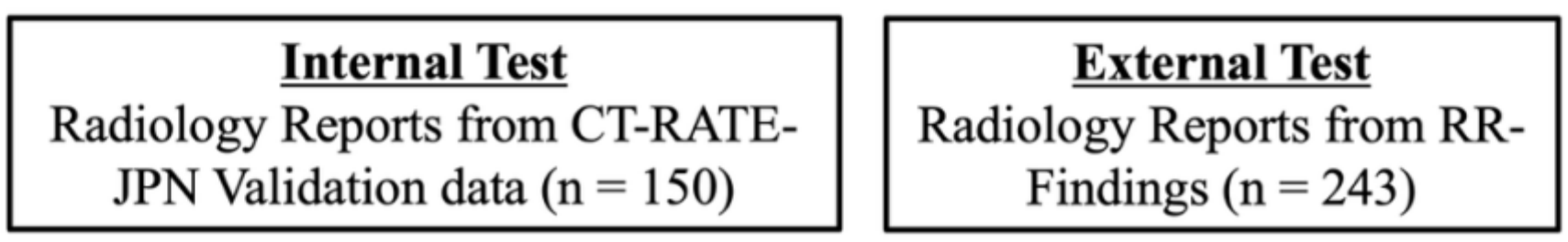

Figure 1: Overview of the CT-RATE-JPN dataset and data splitting approach. A total of 22,778 reports were used for training and validation (split 4:1 into 18,222 training reports and 4,556 validation reports), and 150 radiologist-verified reports served as the original test set. In addition, an external evaluation set of 243 Japanese chest CT reports (RR-Findings), independently annotated with the same 18 CT-RATE findings schema, was incorporated to assess generalization to authentic clinical reporting.

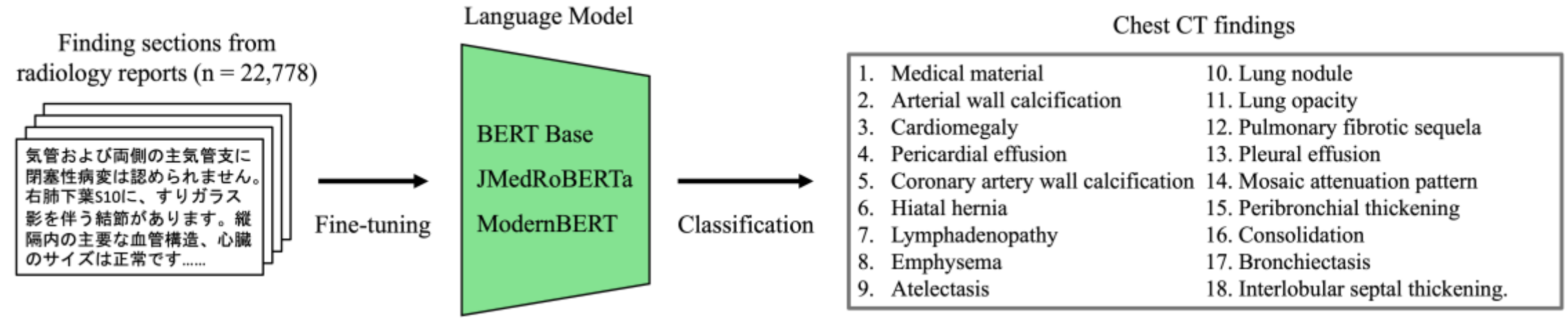

Figure 2: Overview of the chest CT findings classification framework. The process begins with the extraction of finding sections from a large dataset of radiology reports (n = 22,778). These text data are then used to fine-tune three Japanese language models (BERT Base, JMedRoBERTa, and ModernBERT) for chest CT findings classification task.

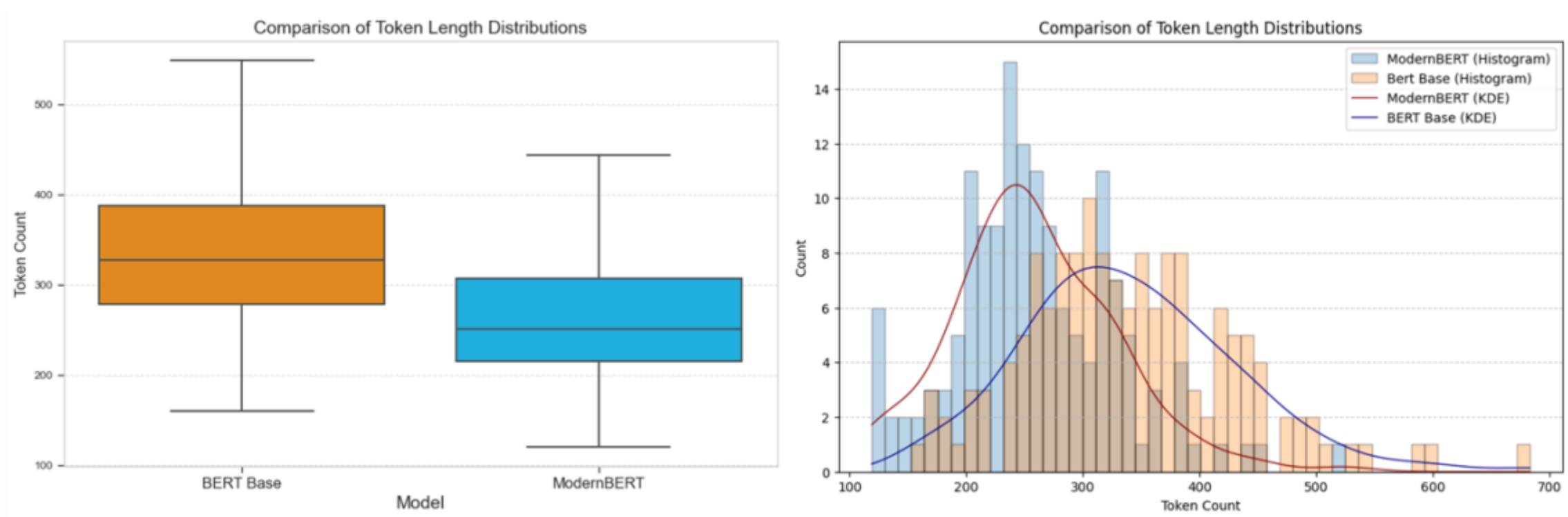

Figure 3: Comparison of token length distributions between ModernBERT and BERT Base on radiological reports of test data. Left: Box plot showing the lower median and narrower distribution for ModernBERT compared to BERT Base. Right: Histogram with density curves illustrating ModernBERT's more efficient tokenization, with consistently lower token counts across the dataset. KDE: Kernel Density Estimation.

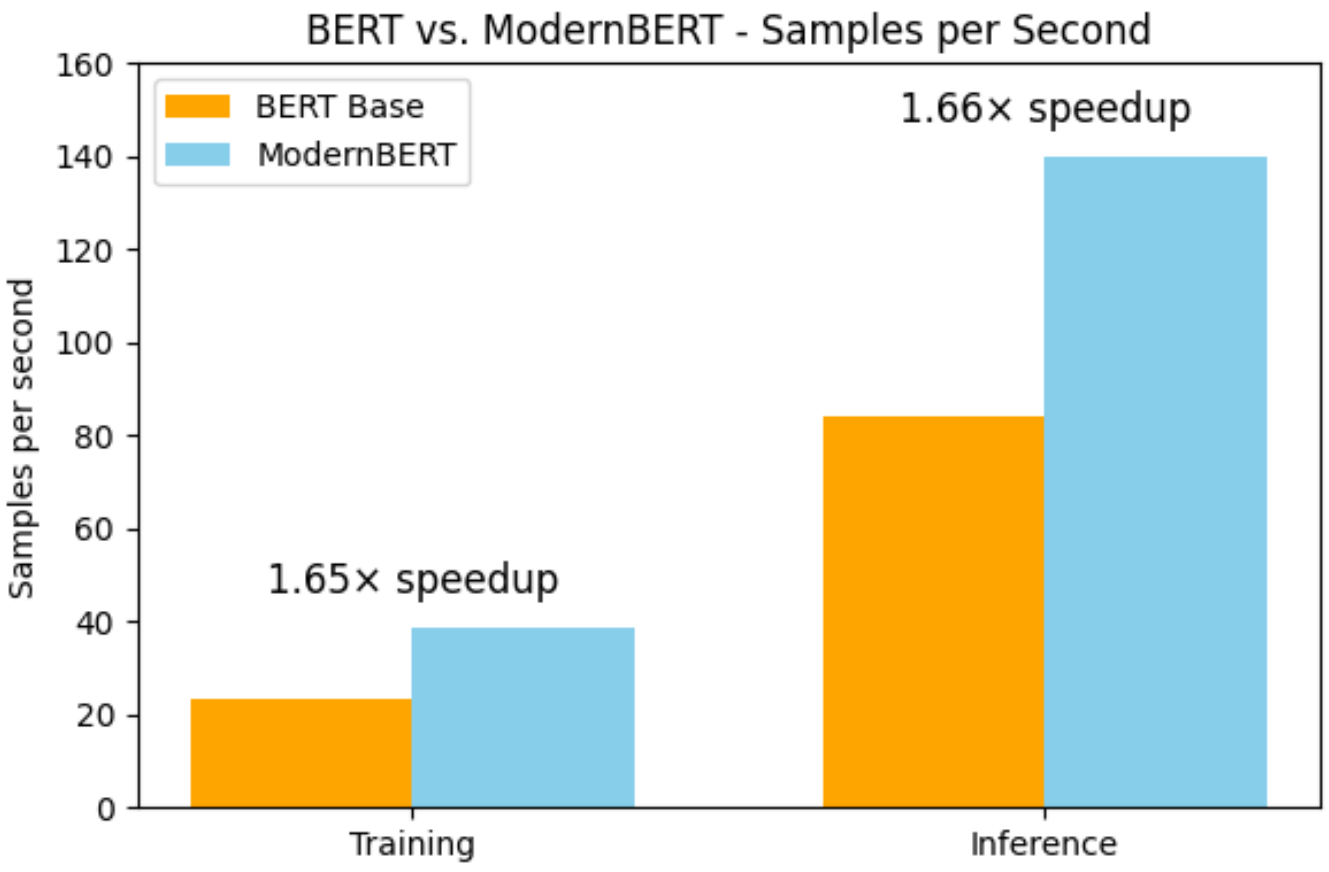

Figure 4: A bar chart comparing the training and inference speeds (in samples per second) of BERT Base and ModernBERT. ModernBERT demonstrates a 1.65× speedup during training and a 1.66× speedup during inference compared to BERT Base.

| findings | train (n = 22778) | valid (n = 150) |
| --- | --- | --- |
| Medical material | 2886 (12.7%) | 14 (9.3%) |
| Arterial wall calcification | 6697 (29.4%) | 49 (32.7%) |
| Cardiomegaly | 2585 (11.3%) | 25 (16.7%) |
| Pericardial effusion | 1702 (7.5%) | 12 (8.0%) |
| Coronary artery wall calcification | 5970 (26.2%) | 45 (30.0%) |
| Hiatal hernia | 3421 (15.0%) | 24 (16.0%) |
| Lymphadenopathy | 6105 (26.8%) | 37 (24.7%) |
| Emphysema | 4668 (20.5%) | 31 (20.7%) |
| Atelectasis | 6166 (27.1%) | 49 (32.7%) |
| Lung nodule | 10856 (47.7%) | 82 (54.7%) |
| Lung opacity | 8831 (38.8%) | 55 (36.7%) |
| Pulmonary fibrotic sequela | 6434 (28.2%) | 47 (31.3%) |
| Pleural effusion | 2858 (12.5%) | 19 (12.7%) |
| Mosaic attenuation pattern | 1794 (7.9%) | 25 (16.7%) |
| Peribronchial thickening | 2496 (11.0%) | 21 (14.0%) |
| Consolidation | 4230 (18.6%) | 24 (16.0%) |
| Bronchiectasis | 2402 (10.5%) | 20 (13.3%) |
| Interlobular septal thickening | 1902 (8.4%) | 7 (4.7%) |

Supplementary Table S1: Distribution of the 18 medical conditions and findings in the CT-RATE-JPN dataset across training (n = 22,778) and test (n = 150) sets. Values represent the number of occurrences and corresponding percentages for each condition.

| Findings | BERT Base | JMedRoBERTa | ModernBERT |
| --- | --- | --- | --- |
| Medical material | 0.9901 | 1.0000 | 0.9841 |
| Arterial wall calcification | 1.0000 | 0.9996 | 1.0000 |
| Cardiomegaly | 0.9957 | 0.9720 | 1.0000 |
| Pericardial effusion | 1.0000 | 1.0000 | 0.9693 |
| Coronary artery wall calcification | 0.9995 | 0.9991 | 0.9991 |
| Hiatal hernia | 1.0000 | 1.0000 | 1.0000 |
| Lymphadenopathy | 0.9875 | 0.9907 | 0.9815 |
| Emphysema | 0.9603 | 0.9826 | 0.9970 |
| Atelectasis | 1.0000 | 1.0000 | 1.0000 |
| Lung nodule | 0.9937 | 0.9895 | 0.9925 |
| Lung opacity | 0.9848 | 0.9851 | 0.9917 |
| Pulmonary fibrotic sequela | 0.9725 | 0.9769 | 0.9482 |
| Pleural effusion | 0.9974 | 0.9572 | 0.9223 |
| Mosaic attenuation pattern | 1.0000 | 1.0000 | 1.0000 |
| Peribronchial thickening | 0.9480 | 0.9437 | 0.9567 |
| Consolidation | 0.9851 | 0.9241 | 0.9521 |
| Bronchiectasis | 0.8732 | 0.9687 | 0.9925 |
| Interlobular septal thickening | 1.0000 | 1.0000 | 0.9821 |
| mAP | 0.9827 | 0.9827 | 0.9816 |

Supplementary Table S2: Comparison of F1 scores across BERT Base, JMedRoBERTa, and ModernBERT for classification of 18 medical findings in chest CT reports on the internal test dataset. Average Precision (AP) is reported as mean AP (mAP) across all findings.

| Findings | BERT Base | JMedRoBERTa | ModernBERT |
|---|---|---|---|
| Lymphadenopathy | 0.9844 | 0.9872 | 0.9912 |
| Lung nodule | 0.9663 | 0.9822 | 0.9397 |
| Pleural effusion | 1.0000 | 1.0000 | 1.0000 |
| Lung opacity | 0.9845 | 0.9687 | 0.9595 |
| Emphysema | 0.9893 | 0.9908 | 0.9864 |
| Atelectasis | 1.0000 | 0.9861 | 0.9710 |
| Pulmonary fibrotic sequela | 0.6748 | 0.6236 | 0.6346 |
| Consolidation | 0.7101 | 0.8748 | 0.8626 |
| Interlobular septal thickening | 1.0000 | 0.9861 | 1.0000 |
| Pericardial effusion | 1.0000 | 0.8088 | 1.0000 |
| Cardiomegaly | 0.6078 | 0.2795 | 0.3138 |
| Arterial wall calcification | 0.2500 | 0.5000 | 0.5000 |
| Bronchiectasis | 1.0000 | 1.0000 | 1.0000 |
| Coronary artery wall calcification | 1.0000 | 1.0000 | 0.5000 |
| Peribronchial thickening | 0.0119 | 0.1111 | 0.0053 |
| mAP | 0.8119 | 0.8066 | 0.7776 |

Supplementary Table S3: Comparison of F1 scores across BERT Base, JMedRoBERTa, and ModernBERT for classification of 15 medical findings in chest CT reports on the external test dataset. Average Precision (AP) is reported as mean AP (mAP) across all findings.

**Original reports**

気管と主気管支は開いています。縦隔
に病的なリンパ節は認められませんで
した。心臓および縦隔の血管構造は正
常な形態を示しています。食道は正常
範囲内です。両側胸腔において胸水や
胸膜肥厚は認められませんでした。両
肺の実質の評価では、斑状の辺縁胸膜
下のすりガラス影によるcrazy paving
appearanceの所見が両肺に見られま
した。ウイルス性肺炎の可能性があり
ますか？CTの関連スコアは軽度と評
価されました。上腹部の断面では、両
側副腎は正常な形態を示しています。
腹部の断面で有意な病変は認められま
せんでした。骨構造に明らかな病変は
認められませんでした。

**Tokenized by BERT**

[CLS] / 気管 / と / 主 / ##気 / 管 / 支 / は / 開い /
て / い / ます / 。 / 縦 / ##隔 / に / 病 / ##的 / な /
リンパ / 節 / は / 認め / られ / ませ / ん / でし / た
/ 。 / 心臓 / および / 縦 / ##隔 / の / 血管 / 構造 /
は / 正常 / な / 形態 / を / 示し / て / い / ます / 。
/ 食 / ##道 / は / 正常 / 範囲 / 内 / です / 。 / 両側
/ 胸 / ##腔 / に / おい / て / 胸 / ##水 / や / 胸 /
##膜 / 肥 / ##厚 / は / 認め / られ / ませ / ん / で
し / た / 。 / 両 / ##肺 / の / 実質 / の / 評価 / で /
は / 、 / 斑 / ##状 / の / 辺 / ##縁 / 胸 / ##膜 / 下
/ の / すり / ガラス / 影 / に / よる / c / ##ra /
##zy / p / ##av / ##ing / a / ##pp / ##ear /
##ance / の / 所 / ##見 / が / 両 / ##肺 / に / 見 /
られ / まし / た / 。 / ウイルス / 性 / 肺炎 / の / 可
能 / 性 / が / あり / ます / か / ? / CT / の / 関連 /
スコア / は / 軽 / ##度 / と / 評価 / さ / れ / まし /
た / 。 / 上 / 腹部 / の / 断面 / で / は / 、 / 両側 /
副 / ##腎 / は / 正常 / な / 形態 / を / 示し / て / い
/ ます / 。 / 腹部 / の / 断面 / で / 有 / ##意 / な /
病 / ##変 / は / 認め / られ / ませ / ん / でし / た / 。
/ 骨 / 構造 / に / 明らか / な / 病 / ##変 / は / 認め
/ られ / ませ / ん / でし / た / 。 / [SEP]

**Tokenized by ModernBERT**

<s> / 気 / 管 / と / 主 / 気管支 / は / 開いて / いま
す / 。 / 縦 / 隔 / に / 病 / 的な / リンパ節 / は / 認
められ / ませんでした / 。 / 心臓 / および / 縦 / 隔
/ の / 血管 / 構造 / は / 正常な / 形態 / を示してい
ます / 。 / 食道 / は / 正常 / 範囲内 / です / 。 / 両
側 / 胸 / 腔 / において / 胸 / 水 / や / 胸 / 膜 / 肥 /
厚 / は / 認められ / ませんでした / 。 / 両 / 肺 / の
/ 実質 / の評価 / では / 、 / 斑 / 状の / 辺 / 縁 / 胸
/ 膜 / 下の / すり / ガラス / 影 / による / crazy /
▁paving / ▁appear / ance / の / 所見 / が / 両 /
肺 / に / 見られました / 。 / ウイルス / 性 / 肺炎 /
の / 可能性があります / か / ？ / CT / の関連 / スコ
ア / は / 軽度 / と評価 / されました / 。 / 上 / 腹部
/ の / 断面 / では / 、 / 両側 / 副 / 腎 / は / 正常な
/ 形態 / を示しています / 。 / 腹部 / の / 断面 / で
/ 有意 / な / 病変 / は / 認められ / ませんでした / 。
/ 骨 / 構造 / に / 明らかな / 病変 / は / 認められ /
ませんでした / 。 / </s>

Supplementary Figure S1: Comparison of tokenization between BERT and ModernBERT models on a complete medical report. The figure presents the original report text (left column) alongside its tokenized representations by BERT (middle column) and ModernBERT (right column). Tokens are separated by "/" symbols, with "##" in the BERT column indicating subtokens.

- **Term: 主気管支 (Main Bronchi)**

  *ModernBERT:* 「主」,「気管支」 → **2 tokens**
  *BERT:* 「主」,「##気」,「管」,「支」 → **4 tokens**

- **Term: リンパ節 (Lymph Nodes)**

  *ModernBERT:* 「リンパ節」 → **1 token**
  *BERT:* 「リンパ」,「節」 → **2 tokens**

- **Phrase: 認められませんでした (Was Not Observed)**

  *ModernBERT:* 「認められ」,「ませんでした」 → **2 tokens**
  *BERT:* 「認め」,「られ」,「ませ」,「ん」,「でし」,「た」 → **6 tokens**

- **Phrase: crazy paving appearance**

  *ModernBERT:* “crazy”, “paving”, “appear”, “ance” → **4 tokens**
  *BERT:* "c", "##ra", "##zy", "p", "##av", "##ing", "a", "##pp", "##ear", "##ance” → **10 tokens**

Supplementary Figure S2: Tokenization comparison of representative Japanese medical terms and a mixed English phrase using ModernBERT and BERT. The “##” symbol in BERT indicates subword tokenization, which splits words into smaller fragments. In contrast, ModernBERT captures these terms more efficiently, using fewer tokens while preserving their original meaning.

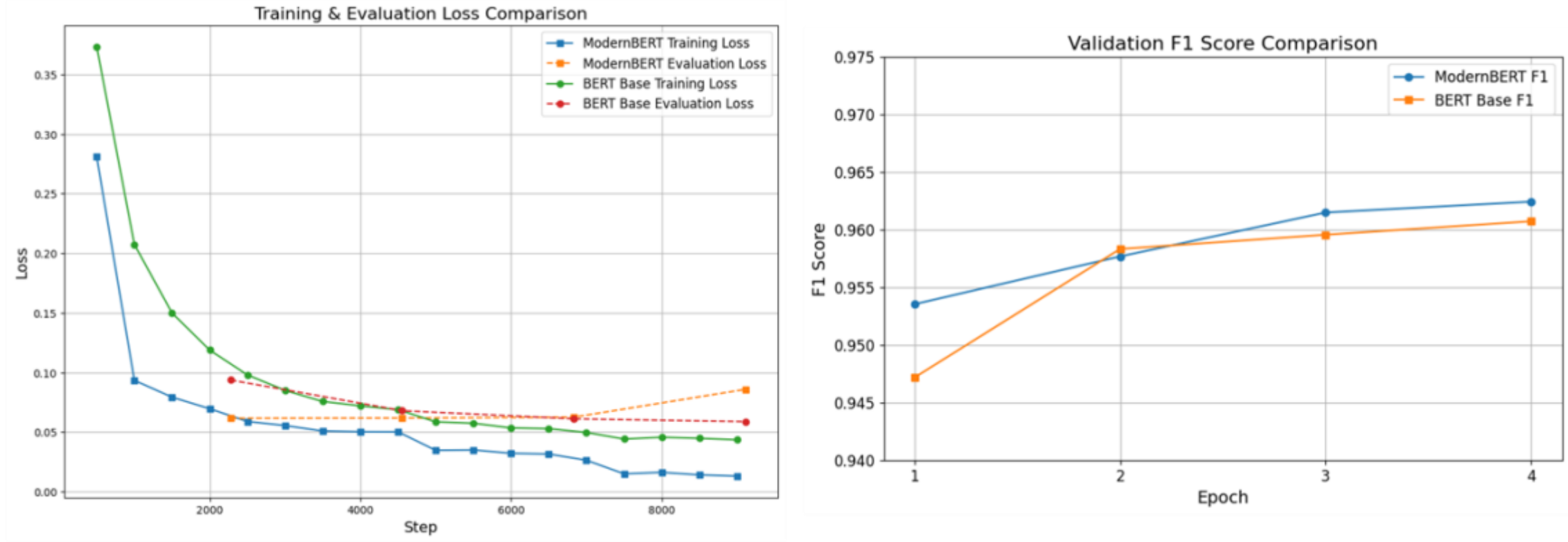

Supplementary Figure S3: (Left) Training and evaluation loss curves for BERT Base and ModernBERT across training steps, with training losses shown as solid lines and evaluation losses as dashed lines. (Right) Validation F1 scores by epoch, comparing performance of BERT Base and ModernBERT.

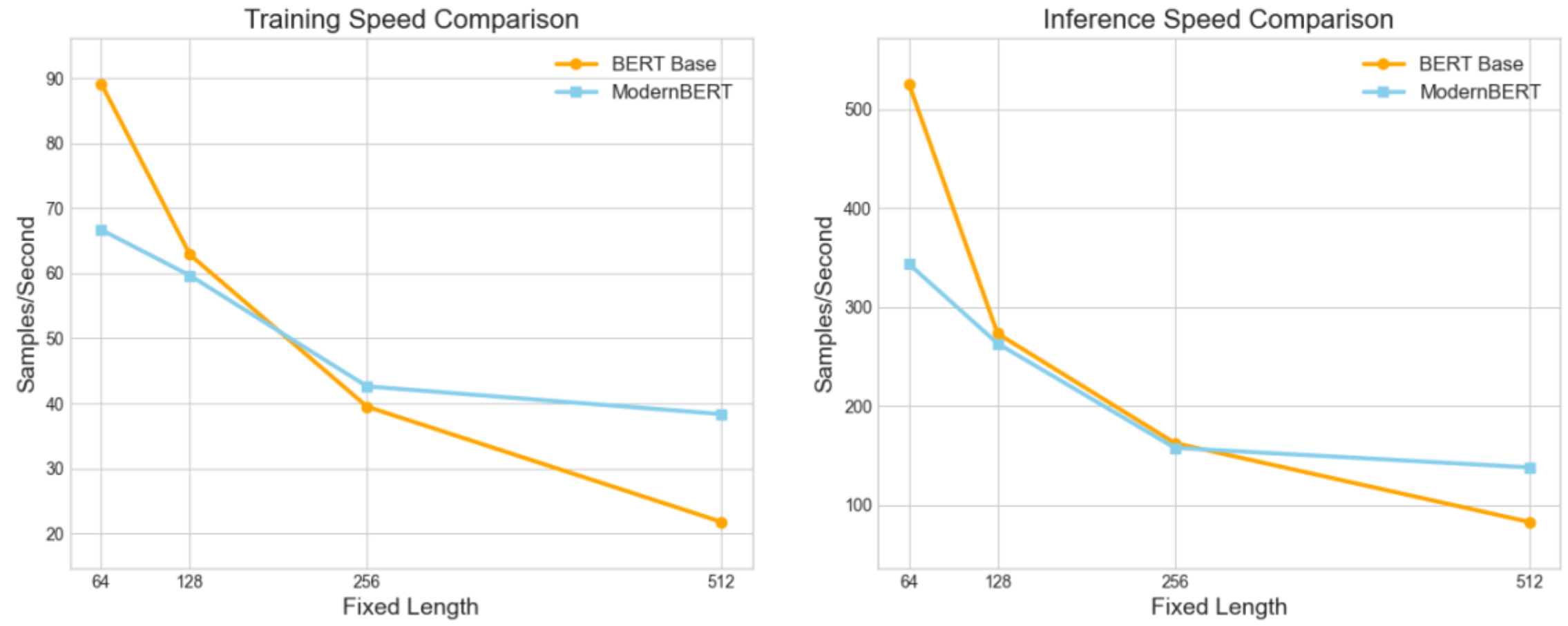

Supplementary Figure S4: Comparison of processing efficiency between BERT Base and ModernBERT across different fixed token lengths (64, 128, 256, 512). Left: Training samples processed per second. Right: Inference samples processed per second.